\documentclass{article}
\usepackage{iclr2027_conference,times}

\usepackage{amsmath,amsfonts,bm}

\def\eqref#1{equation~\ref{#1}}

\def\1{\bm{1}}

\DeclareMathAlphabet{\mathsfit}{\encodingdefault}{\sfdefault}{m}{sl}
\SetMathAlphabet{\mathsfit}{bold}{\encodingdefault}{\sfdefault}{bx}{n}

\usepackage{hyperref}
\usepackage{url}
\usepackage{algorithm}
\usepackage{algorithmic}
\usepackage{graphicx}
\usepackage{booktabs}
\usepackage{flafter}
\usepackage{wrapfig}
\usepackage{placeins}
\usepackage{needspace}

\title{Learning Quality-Controllable Robot Policies with Iterative Hindsight Relabeling}

\author{%
\normalfont
\textbf{Sinjae Kang}$^{1}$\thanks{Corresponding author: Sinjae Kang (\texttt{sinjae.kang1@gmail.com}).}%
\quad
\textbf{Dong Kyu Shin}$^{1}$
\quad
\textbf{Jaehong Kim}$^{2}$
\quad
\textbf{Kamin Lee}$^{2}$
\quad
\textbf{Hyoeun Kim}$^{2}$
\\[0.3em]
\textbf{Kaixin Wang}$^{3}$
\quad
\textbf{Li Zhao}$^{3}$
\quad
\textbf{Seonghyeon Ye}$^{1\dagger}$
\quad
\textbf{Kimin Lee}$^{1,4\dagger}$
\\[0.8em]
\normalfont
$^{1}$KAIST
\quad
$^{2}$LGE
\quad
$^{3}$Microsoft Research
\quad
$^{4}$Config
}
\iclrfinalcopy

\begin{document}
\raggedbottom
\maketitle\pagestyle{plain}\thispagestyle{plain}

\begin{abstract}
Robot policy evaluation often relies on a binary success signal for task completion. In real-world deployment, however, desired execution behaviors are also defined by quality attributes such as execution speed and motion smoothness, which need to be dynamically controlled according to task requirements. To address this, we introduce Quality-Controllable Policy (QCP), a robot policy that enables inference-time control over both individual quality attributes and their combinations. To extend the controllable quality range beyond initial demonstrations, we further propose an iterative hindsight relabeling loop: QCP collects online rollouts under target quality labels, adds successful trajectories to the dataset, relabels the accumulated trajectories in hindsight according to their measured quality, and fine-tunes the policy. Experiments on Robomimic and RoboEval demonstrate that QCP consistently improves both task execution quality and success rates across iterations, producing successful rollouts that exceed the quality of the initial demonstrations. Finally, we validate QCP on real-world manipulation tasks, where controlling speed increases throughput in conveyor operations and controlling transport stability improves success when carrying wine glasses, demonstrating that quality-controllable policies yield practical gains in physical deployment.\par\vspace{4pt}\noindent\textbf{Project page:} \url{https://sinnnj.github.io/qcp}
\end{abstract}
\section{Introduction}
\label{sec:introduction}

Robot foundation models have shown strong generalization across tasks, environments, and objects \citep{black2025pi05,bjorck2025grootn15,ye2026dreamzero}.
These policies are usually trained with behavior cloning (BC) on successful demonstrations, which treats every successful trajectory as equally desirable regardless of how the task was executed.
As a result, the learned policy reproduces the mixture of execution behaviors in the data, whether fast or slow, smooth or jerky, and offers no way to steer toward the behavior a deployment actually requires.
In practice, however, the desired behavior of a robot is not defined by task success alone \citep{wang2025roboeval,liu2026evalactions,mao2026beyondsuccess} and different tasks call for different \textit{quality attributes}.
For example, a robot deployed on a conveyor line should prioritize execution speed, whereas a robot carrying wine glasses on a tray must prioritize transport stability.

To address this, prior work has proposed quality-controllable policies whose execution behavior can be adjusted along a chosen quality axis at inference time \citep{zhan2020calibratable,yang2025bcpmi,petitbois2026sciql}.
However, such policies are trained on static demonstration datasets, which typically cover only a bounded quality range and lack the high-performance regimes needed in deployment (e.g., maximum speed or ideal smoothness).
Our goal is therefore not only to learn a quality-controllable policy, but also to continuously expand its controllable range beyond the initial demonstrations through online data collection and policy updates.



\begin{figure}[!t]
    \centering
    \setlength{\parskip}{0pt}
    \setlength{\abovecaptionskip}{8pt}
    \includegraphics[width=0.95\linewidth]{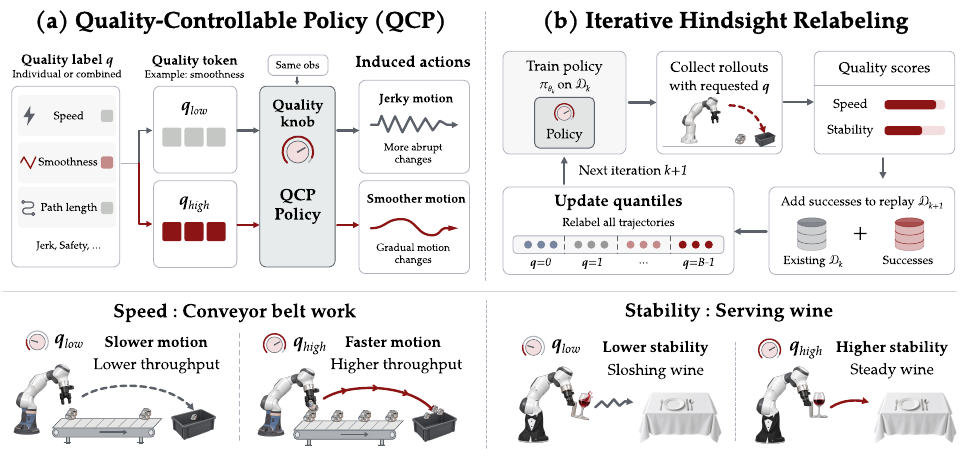}
    \par\nointerlineskip
    \caption{\textbf{Overview of QCP and iterative hindsight relabeling.} \textit{Left:} QCP controls execution quality at inference time through conditioning on quantized quality tokens. \textit{Right:} Iterative hindsight relabeling shifts the quality distribution of successful rollouts toward higher quality, expanding the controllable range beyond the initial demonstrations.}
    \label{fig:hqr_overview}
\end{figure}

In this paper, we introduce \emph{Quality-Controllable Policy (QCP)}, which addresses this limitation of BC by making execution quality an explicit conditioning input (Figure~\ref{fig:hqr_overview}).
For each demonstration, we measure the target quality attributes with rule-based score functions and quantize the scores into discrete quality labels.
The policy is then trained to generate actions conditioned on both the observation history and these labels.
Rather than averaging over execution behaviors, QCP learns task completion from all successful trajectories while distinguishing how each one was executed.
At inference time, the user specifies a target label to steer execution behavior, for example, requesting the highest speed label for conveyor operations.

To extend the controllable quality range beyond the demonstrations, we pair QCP with \emph{iterative hindsight relabeling}.
At each iteration, we collect rollouts conditioned on the highest quality label, keep the successful ones, relabel them by the quality they actually achieved, and add them to the dataset.
We then recompute the quality labels of all trajectories and retrain the policy.
As higher-quality rollouts accumulate, the highest label comes to represent progressively higher quality, allowing the policy to reach operating regimes absent from the initial demonstrations.
Because quality is measured by rule-based score functions, this loop requires no manual quality annotation.

Experiments on Robomimic~\citep{mandlekar2022robomimic} and RoboEval~\citep{wang2025roboeval} show that QCP with iterative hindsight relabeling improves both target quality and task success across iterations, outperforming quality-filtered BC and RL baselines~\citep{mao2024sorl,petitbois2026sciql,wagenmaker2025dsrl,dong2025expo}.
Iterative relabeling pushes the quality distribution well beyond the range of the initial demonstrations, reaching an IQR-normalized Wasserstein distance above 1.5.
Beyond single-attribute control, QCP controls multiple attributes jointly, improving six quality attributes simultaneously in RoboEval.
Finally, in real-world experiments, controlling execution speed substantially increases throughput in conveyor operations, and controlling transport stability improves success when carrying wine glasses on a tray.






\section{Related Work}
\label{sec:related_work}

\paragraph{Policy learning with quality control.}
\label{sec:related_stylized_policy_learning}

Prior work studies how to learn different ways of executing a task, representing behavioral styles through learned trajectory representations or explicit labels \citep{petitbois2025wzbc,zhan2020calibratable,yang2025bcpmi}. Learning a requested style, however, can conflict with task performance. SORL addresses this tension by combining trajectory clustering with task advantage weighting \citep{mao2024sorl}, while SCIQL combines task and style advantages to account for both task performance and style alignment \citep{petitbois2026sciql}. QCP combines quantized quality conditioning with iterative hindsight relabeling. Our experiments show simultaneous gains in execution quality and task success, demonstrating that this trade-off can be overcome.

\paragraph{Online improvement of behavioral quality.}
\label{sec:related_online_quality_improvement}

Freeform Preference Learning (FPL) learns a reward model for the desired quality axes from human comparisons of robot trajectories and trains policies conditioned on the predicted rewards \citep{torne2026fpl}. This supports online improvement through repeated rollout collection and policy training, but requires additional human comparisons at each round. In text generation, Quark repeatedly samples under the highest reward token, relabels accumulated samples by reward quantiles, and retrains the language model conditioned on the assigned labels \citep{lu2022quark}. QCP applies a similar loop to robot execution quality, using automatically computed quality labels and accumulated successful trajectories to extend the controllable range beyond the initial demonstrations without additional human annotation.

\section{Method}
\label{sec:method}

In real-world deployment, users often care about quality metrics beyond binary task success, such as execution speed or motion smoothness.
To account for these, we propose Quality-Controllable Policy (QCP), which adjusts its execution behavior along user-specified quality axes by conditioning on quantized quality levels (Section~\ref{sec:quality_definition}).
Because demonstration data typically covers only a limited range of quality, we extend QCP's controllable quality range through iterative hindsight relabeling, alternating between collecting online rollouts, relabeling trajectories by their achieved quality, and updating the policy (Section~\ref{sec:quality_replay}).
Figure~\ref{fig:hqr_overview} illustrates the overall framework.

\subsection{Quality-Controllable Policy}
\label{sec:quality_definition}

\paragraph{Quality scores and quantization.}
Let $\mathcal{D}$ be a dataset of successful demonstrations, where each trajectory $\tau=(o_0,a_0,\ldots,o_{T-1},a_{T-1},o_T)$ consists of observations $o_t$ and executed actions $a_t$, with $T$ the number of executed actions. We assume access to a rule-based score function $m(\tau)\in\mathbb{R}$ that measures a quality axis of interest.
The simplest example is $m(\tau)=\frac{1}{T}$ as a proxy for execution speed, and our experiments additionally consider more complex scores such as normalized action differences for smoothness and total joint path length for efficiency (see Appendix~\ref{app:quality_score_functions} for detailed definitions).

We then quantize the scores into $B$ equal-frequency bins and use the bin index as a discrete quality label:
\begin{equation}
q(\tau)=\bigl\lceil B\cdot\hat F(m(\tau))\bigr\rceil-1\in\{0,\ldots,B-1\},
\qquad
\hat F(s)=\frac{1}{|\mathcal{D}|}\sum_{\tau'\in\mathcal{D}}\mathbf{1}\{m(\tau')\le s\},
\label{eq:quality_label}
\end{equation}
where $\hat F$ is the empirical CDF of scores in $\mathcal{D}$, so that $q(\tau)=j$ if and only if $m(\tau)$ lies between the $j/B$ and $(j+1)/B$ empirical quantiles. Every trajectory in $\mathcal{D}$ is thus paired with its label $q(\tau)$, and we train a quality-conditioned policy on this labeled dataset.\footnote{See Section~\ref{sec:analysis_discretization} for a comparison with continuous score conditioning, and Appendix~\ref{sec:analysis_subtrajectory} for a comparison with assigning a separate quality label to shorter segments of each trajectory.}

\paragraph{Quality-conditioned policy training.}
We train a policy $\pi_\theta$ that, given a history of $L$ observations $h_t=(o_{t-L+1},\ldots,o_t)$ and a quality label $q$, predicts a chunk of $H$ future actions $A_t=(a_t,\ldots,a_{t+H-1})$.
We construct training samples $(h_t,A_t,q)$ by slicing each trajectory $\tau\in\mathcal{D}$ at every time step $t$ and attaching its label $q=q(\tau)$.
Denoting the resulting set of training samples by $\mathcal{D}_{\mathrm{tr}}$, the policy is trained with the following quality-conditioned behavior cloning objective:
\begin{equation}
\mathcal{L}(\theta) = -\mathbb{E}_{(h_t,A_t,q)\sim\mathcal{D}_{\mathrm{tr}}}
\left[\log\pi_\theta(A_t\mid h_t,q)\right].
\label{eq:quality_bc_loss}
\end{equation}

We model $\pi_\theta$ as a flow matching policy~\citep{lipman2023flowmatching}, widely used in recent robot policies~\citep{black2025pi05,bjorck2025grootn15,kang2026warmprior}, which optimizes~\eqref{eq:quality_bc_loss} in practice by training a velocity field with a regression objective.
Given noise $\epsilon\sim\mathcal{N}(0,I)$ and flow time $u\sim\mathcal{U}[0,1]$, we form the interpolated action chunk $A_t^{u}=(1-u)\epsilon+uA_t$ and train a velocity field $v_\theta$ to predict the target $A_t-\epsilon$:
\begin{equation}
\mathcal{L}_{\mathrm{FM}}(\theta)= \mathbb{E}_{(h_t,A_t,q)\sim\mathcal{D}_{\mathrm{tr}},\;u,\;\epsilon}
\left[
\left\|v_\theta\bigl(A_t^{u},u,h_t,q\bigr)-(A_t-\epsilon)\right\|^2
\right],
\label{eq:quality_flow_loss}
\end{equation}
where the norm is taken over all action dimensions and time steps in the chunk.
We consider two cases for conditioning on quality $q$. For simple policies trained from scratch, an embedding layer maps $q$ to a learned vector $\mathbf{e}_q\in\mathbb{R}^d$, which is concatenated with the encoded observation history to condition the velocity network.\footnote{For joint control of multiple quality attributes, we compute a separate quality label for each attribute, map each label through its own embedding layer, and let the embeddings jointly condition the velocity network.} For vision-language-action (VLA) models such as $\pi_{0.5}$~\citep{black2025pi05}, we instead insert the label into the language prompt as text (e.g., ``quality level: 3''), similar to $\pi_{0.7}$ \citep{intelligence2026pi07}, and fine-tune the model with its original training recipe. This enables the model to interpret quality levels through language conditioning without architectural changes.

\paragraph{Inference during deployment.}
At inference time, the quality label becomes a control knob: the user specifies a target bin $q^\star\in\{0,\ldots,B-1\}$, and actions are generated by integrating the learned velocity field $v_\theta(\cdot,u,h_t,q^\star)$ from $u=0$ to $u=1$ starting from Gaussian noise. We find that QCP responds reliably to this input, shifting its behavior toward the requested bin and attaining a quality score within the corresponding range (Figure~\ref{fig:quality_control}).
Unless noted, we condition on the highest bin $q^\star=B-1$ throughout our evaluation and in the iterative training procedure of Section~\ref{sec:quality_replay}.

\subsection{Iterative Hindsight Relabeling}
\label{sec:quality_replay}

\begingroup
\begin{wrapfigure}{r}{0.42\linewidth}
\begin{minipage}{\linewidth}
\begin{algorithm}[H]
\small
\caption{QCP with Iterative Hindsight Relabeling}
\label{alg:qcp}
\begin{algorithmic}[1]
\REQUIRE Demonstrations $\mathcal{D}$, score $m$, $B$ bins, $K$ collection rounds.
\STATE Set $\mathcal{D}_0\gets\mathcal{D}$, $q^\star\gets B-1$.
\STATE Label $\mathcal{D}_0$ and train $\pi_{\theta_0}$ as in Section~\ref{sec:quality_definition}.
\FOR{$k=0,\ldots,K-1$}
    \STATE Collect rollouts with $\pi_{\theta_k}(\cdot\mid h_t,q^\star)$.
    \STATE Compute $m(\tau)$ for each rollout.
    \STATE Append successful rollouts to $\mathcal{D}_k$ to obtain $\mathcal{D}_{k+1}$.
    \STATE Compute $\hat F$ from scores in $\mathcal{D}_{k+1}$.
    \STATE Recompute $q(\tau)$ for all $\tau\in\mathcal{D}_{k+1}$ using Equation~\ref{eq:quality_label}.
    \STATE Train $\pi_{\theta_{k+1}}$ from scratch on $\mathcal{D}_{k+1}$ using Equation~\ref{eq:quality_flow_loss}.
\ENDFOR
\end{algorithmic}
\end{algorithm}
\end{minipage}
\end{wrapfigure}

A policy trained only on demonstrations is largely bounded by the quality present in $\mathcal{D}$.
To push beyond this range, we introduce a simple iterative pipeline that alternates between rollout collection, hindsight relabeling, and policy updates.
The key observation is that rollouts conditioned on the highest bin can exceed the best demonstrations (Figure~\ref{fig:hqr_distribution_evolution}).
We attribute this to two factors.
First, the label is attached to every action chunk of a trajectory, so the policy learns how each stage of the task is executed in the high-quality demonstrations and can recombine these segments into a trajectory whose quality exceeds that of any single demonstration.
Second, action generation is stochastic, and conditioning on the highest bin of the target quality axis raises the probability that a sampled rollout attains higher quality along that axis than the demonstrations.
Folding these rollouts into the dataset and recomputing the labels moves the highest bin upward, and retraining on the relabeled data lets the policy produce behavior that the previous policy did not; repeating this expands the controllable quality range over iterations (Section~\ref{sec:analysis_conditioning}).

We initialize the dataset as $\mathcal{D}_0=\mathcal{D}$ and train an initial policy $\pi_{\theta_0}$ on it as in Section~\ref{sec:quality_definition}.
At each iteration $k$, we execute $\pi_{\theta_k}$ conditioned on the highest bin $q^\star=B-1$ to collect a batch of rollouts, and keep only those that succeed at the task.\footnote{We also tried including failed rollouts as additional low-quality data, but found that this degraded performance.}
Although every rollout is requested at the highest bin, the realized scores vary.
We therefore relabel each successful rollout in hindsight by its realized score $m(\tau)$ rather than the requested label, and append it to $\mathcal{D}_k$ to obtain $\mathcal{D}_{k+1}$.
We then recompute the quality labels $q(\tau)$ for all trajectories in $\mathcal{D}_{k+1}$, including the original demonstrations, by applying~\eqref{eq:quality_label} with the empirical CDF taken over $\mathcal{D}_{k+1}$.
As higher-scoring rollouts accumulate, the bin boundaries shift upward, progressively raising the quality represented by the highest bin while the original demonstrations preserve the low-quality range and thus widen the controllable range.

Finally, we train $\pi_{\theta_{k+1}}$ on the relabeled $\mathcal{D}_{k+1}$ as in Section~\ref{sec:quality_definition}.
Rather than fine-tuning $\pi_{\theta_k}$, we reinitialize the policy at every iteration, from random weights for policies trained from scratch and from the pretrained checkpoint for VLAs.
This is because relabeling changes which quality range each label refers to, so the label-to-behavior mapping learned in the previous iteration no longer matches the new targets and would need to be unlearned.\footnote{Reinitialization may also help avoid the loss of plasticity and primacy bias observed when networks are trained continually on a growing dataset in deep RL~\citep{nikishin2022primacy}, although we do not study this effect directly.}
We repeat rollout collection, relabeling, and retraining for a fixed number of iterations, as summarized in Algorithm~\ref{alg:qcp}.

\section{Experiments}
\label{sec:experiments}
\begin{figure}[!t]
    \centering
    \setlength{\parskip}{0pt}
    \setlength{\abovecaptionskip}{8pt}
    \setlength{\belowcaptionskip}{0pt}
    \includegraphics[width=\linewidth]{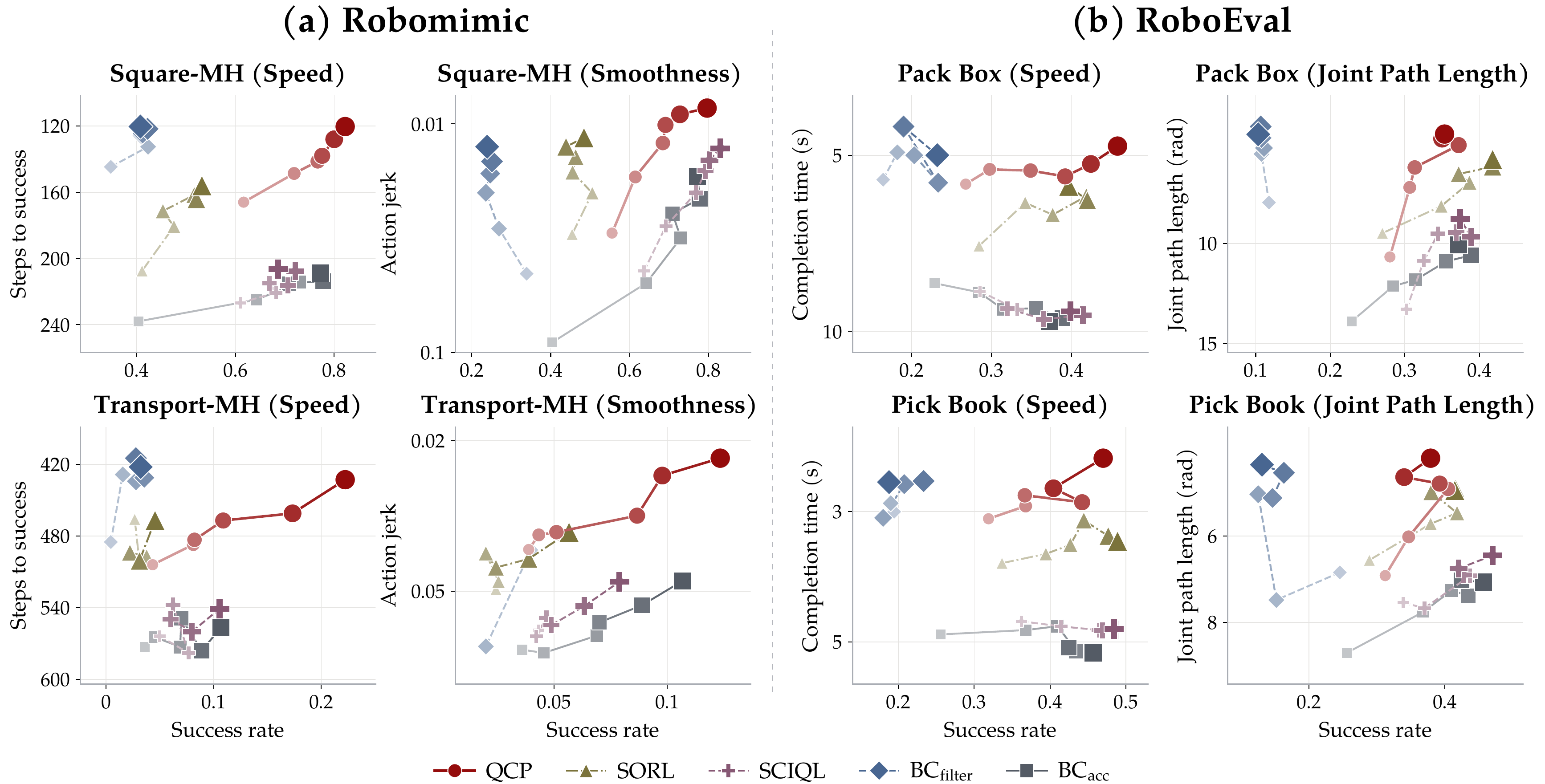}
    \par\nointerlineskip
    \caption{Comparison of QCP with BC baselines (BC$_\text{filter}$, BC$_\text{acc}$) and behavioral quality learning methods (SORL, SCIQL) across iterations on (a) Robomimic and (b) RoboEval. Each plot shows task success rate against a quality metric, with the y-axis oriented so that higher is better. Darker and larger markers correspond to later iterations. QCP improves quality and task success together more effectively than the baselines. Results are averaged over three random seeds.}
    \label{fig:hqr_main_results}
\end{figure}

We investigate the performance of QCP with iterative hindsight relabeling by focusing on the following four research questions:
\par\begingroup
\setlength{\leftskip}{0.7em}
\setlength{\parskip}{4pt}
\noindent (1) Does iterative hindsight relabeling consistently improve the target quality of QCP while maintaining task success? (\hyperref[sec:iterative_results]{Section~\ref*{sec:iterative_results}})\par
\noindent (2) Does iterative hindsight relabeling expand the controllable quality range of QCP? (\hyperref[sec:quality_control_results]{Section~\ref*{sec:quality_control_results}})\par
\noindent (3) Can QCP jointly control and improve multiple quality attributes? (\hyperref[sec:multi_quality_results]{Section~\ref*{sec:multi_quality_results}})\par
\noindent (4) Does our approach remain effective on real-world robot tasks? (\hyperref[sec:generalization_results]{Section~\ref*{sec:generalization_results}})\par
\endgroup

\subsection{Experimental setup}
\label{sec:experimental_setup}

\paragraph{Tasks and datasets.}
Our main simulation experiments use Square and Transport tasks from Robomimic \citep{mandlekar2022robomimic} and Pack Box and Pick Book tasks from RoboEval \citep{wang2025roboeval}.
For Robomimic, we use the Multi-Human (MH) datasets, whose demonstrations are collected by operators of varying proficiency and thus span a wide range of execution quality. For RoboEval, we use the demonstration datasets with randomized object positions and orientations. For the real-world experiments in Section~\ref{sec:generalization_results}, we use a Franka Research 3 arm with the DROID hardware setup~\citep{khazatsky2024droid}, and collect 400 teleoperated demonstrations for the conveyor task and 50 for the wine-glass transport task to form each task's initial dataset $\mathcal{D}_0$.



\paragraph{Baselines.}
We compare QCP against three groups of baselines.
To test whether data accumulation or quality filtering alone can achieve similar gains, we consider BC trained on accumulated successful rollouts (\textsc{BC}$_{\text{acc}}$), and quality-filtered BC using the best 10\% of successful trajectories along the target axis (\textsc{BC}$_{\text{filter}}$).\footnote{In the real-world experiments, \textsc{BC}$_{\text{filter}}$ keeps the best 20\% instead, which gives the baseline a stronger setting given the small initial datasets (Appendix~\ref{app:baseline_details}).}
To compare against existing methods for learning desired behavioral styles, we evaluate SORL~\citep{mao2024sorl} and SCIQL~\citep{petitbois2026sciql}, which learn style-conditioned policies through offline RL.
Finally, to compare quality conditioning against direct optimization of quality rewards, we evaluate DSRL~\citep{wagenmaker2025dsrl} and EXPO~\citep{dong2025expo}, which improve flow matching policies with online RL.
For both, we add a weighted quality reward to the task reward and train them for 720k environment steps, the budget of QCP's 1,800 rollouts (300 per iteration over six iterations) at a 400-step limit.
Reward settings and interaction budgets are detailed in Appendix~\ref{app:baseline_details}.


\paragraph{Quality metrics.}
On Robomimic, we target execution speed, measured by steps to success, and action smoothness, measured by action jerk. On RoboEval, we target execution speed, measured by completion time, and motion efficiency, measured by total joint path length.
We negate or invert qualities for which lower values indicate higher quality to obtain the score $m(\tau)$ used in \eqref{eq:quality_label}.
In real-world experiments, we target execution speed, measured by end-effector speed, and transport stability, measured by the end-effector tilt from vertical. 
Detailed definitions are provided in Appendix~\ref{app:quality_score_functions}.


\paragraph{Training and evaluation.}
In simulation, we use a simple flow matching policy with an MLP backbone, trained for 100k steps on Robomimic and 30k steps on RoboEval with a batch size of 1024.
In real-world experiments, we fine-tune the $\pi_{0.5}$ vision-language-action (VLA) model~\citep{black2025pi05} for 10 epochs with a batch size of 32.
These settings are fixed across iterations.
We use $B=4$ quality bins and always condition QCP on the highest bin $q^\star=3$ during rollout collection.\footnote{Section~\ref{sec:analysis_discretization} evaluates performance with different numbers of quality bins.}
At each iteration, we collect 300 rollouts per task and seed in simulation, and in the real world, as many rollouts as there are demonstrations in the initial dataset $\mathcal{D}_0$.
These rollouts serve both for evaluation and as training data for the next iteration.
Throughout, \emph{iteration $k$} refers to the policy $\pi_{\theta_k}$ trained on $\mathcal{D}_k$ and to the rollouts it collects. Iteration 0 thus refers to the policy trained on the demonstrations alone, and results at iteration $k$ reflect $k$ rounds of rollout collection.
We report mean quality over successful rollouts and success rate over all rollouts, both averaged across three seeded runs of the full iterative procedure.


\subsection{Iterative improvement over quality and success rate}
\label{sec:iterative_results}

Figure~\ref{fig:hqr_main_results} shows execution quality and task success across iterations.
Across tasks and quality attributes, QCP improves both target quality and task success, whereas the baselines tend to improve one at the expense of the other, each for a different reason.
\textsc{BC}$_{\text{acc}}$ has no quality signal and improves task success alone. \textsc{BC}$_{\text{filter}}$ and SORL, which train only on trajectories whose label matches the request, discard most of the successful data, at a cost in task success. SCIQL keeps all trajectories but weights them by the sum of a success advantage and a quality advantage, which creates a trade-off between the two (Appendix~\ref{app:quality_reward_weight}). QCP avoids these problems: all successful trajectories are used by conditioning on their quality labels, so even low-quality trajectories contribute to learning task completion, while the requested label selects high-quality behavior at inference.

QCP also achieves higher final quality than the flow matching RL baselines DSRL and EXPO (Table~\ref{tab:rl_main}), even when we assign high weights to their quality rewards.\footnote{Appendix~\ref{app:quality_reward_weight} compares quality reward weights for DSRL and EXPO.} We attribute this to the fact that these RL methods for flow matching policies are tuned to consume a success signal, and a reward along a new quality axis is not consumed as effectively. QCP instead provides the quality signal in a supervised manner by conditioning the policy on achieved quality labels, achieving high execution quality together with strong task success.


\begin{table}[!htbp]
\centering
\setlength{\parskip}{0pt}
\setlength{\abovecaptionskip}{0pt}
\setlength{\belowcaptionskip}{6pt}
\caption{\textbf{QCP versus RL baselines.} Results at the final iteration (iteration 5, i.e., after five rounds of rollout collection) are averaged across three random seeds. Higher success rates and lower quality metric values are better.}
\label{tab:rl_main}
\begingroup
\setlength{\tabcolsep}{3pt}

\resizebox{\linewidth}{!}{%
\begin{tabular}{@{}l*{16}{c}@{}}
\toprule
 & \multicolumn{2}{c}{\textbf{Sq+Speed}}
 & \multicolumn{2}{c}{\textbf{Sq+Smoothness}}
 & \multicolumn{2}{c}{\textbf{Tr+Speed}}
 & \multicolumn{2}{c}{\textbf{Tr+Smoothness}}
 & \multicolumn{2}{c}{\textbf{Pack+Speed}}
 & \multicolumn{2}{c}{\textbf{Pack+JPL}}
 & \multicolumn{2}{c}{\textbf{Book+Speed}}
 & \multicolumn{2}{c}{\textbf{Book+JPL}} \\
\cmidrule(lr){2-3}
\cmidrule(lr){4-5}
\cmidrule(lr){6-7}
\cmidrule(lr){8-9}
\cmidrule(lr){10-11}
\cmidrule(lr){12-13}
\cmidrule(lr){14-15}
\cmidrule(lr){16-17}
Method
 & \multicolumn{1}{c}{SR}
 & \multicolumn{1}{c}{Speed}
 & \multicolumn{1}{c}{SR}
 & \multicolumn{1}{c}{Smoothness}
 & \multicolumn{1}{c}{SR}
 & \multicolumn{1}{c}{Speed}
 & \multicolumn{1}{c}{SR}
 & \multicolumn{1}{c}{Smoothness}
 & \multicolumn{1}{c}{SR}
 & \multicolumn{1}{c}{Speed}
 & \multicolumn{1}{c}{SR}
 & \multicolumn{1}{c}{JPL}
 & \multicolumn{1}{c}{SR}
 & \multicolumn{1}{c}{Speed}
 & \multicolumn{1}{c}{SR}
 & \multicolumn{1}{c}{JPL} \\
 & \multicolumn{1}{c}{(\%)}
 & \multicolumn{1}{c}{(steps)}
 & \multicolumn{1}{c}{(\%)}
 & \multicolumn{1}{c}{($\times 10^{-3}$)}
 & \multicolumn{1}{c}{(\%)}
 & \multicolumn{1}{c}{(steps)}
 & \multicolumn{1}{c}{(\%)}
 & \multicolumn{1}{c}{($\times 10^{-3}$)}
 & \multicolumn{1}{c}{(\%)}
 & \multicolumn{1}{c}{(s)}
 & \multicolumn{1}{c}{(\%)}
 & \multicolumn{1}{c}{(rad)}
 & \multicolumn{1}{c}{(\%)}
 & \multicolumn{1}{c}{(s)}
 & \multicolumn{1}{c}{(\%)}
 & \multicolumn{1}{c}{(rad)} \\
\midrule
\textbf{QCP (ours)}
 & 82.2 & \textbf{120.3}
 & 79.7 & \textbf{8.5}
 & \textbf{22.2} & \textbf{432.8}
 & \textbf{12.3} & \textbf{22.2}
 & 45.8 & \textbf{4.82}
 & 35.3 & \textbf{6.42}
 & 47.0 & \textbf{2.43}
 & 37.9 & \textbf{4.63} \\
\addlinespace[0.5pt]
SORL
 & 53.2 & 156.0
 & 48.4 & 11.5
 & 4.6 & 466.8
 & 5.7 & 34.9
 & 39.7 & 5.64
 & 41.8 & 7.11
 & \textbf{48.9} & 3.37
 & 41.4 & 5.15 \\
SCIQL
 & 68.7 & 206.5
 & \textbf{83.0} & 12.8
 & 10.6 & 540.8
 & 7.9 & 47.2
 & 39.9 & 9.24
 & 37.4 & 9.06
 & 48.4 & 4.75
 & \textbf{47.0} & 6.40 \\
DSRL
 & 37.0 & 208.8
 & 43.0 & 46.9
 & 4.0 & 549.5
 & 7.0 & 79.4
 & \textbf{60.7} & 6.35
 & \textbf{50.7} & 66.05
 & 0.0 & ---
 & 0.0 & --- \\
EXPO
 & \textbf{96.0} & 162.5
 & 79.0 & 157.4
 & 11.0 & 567.7
 & 3.0 & 176.5
 & 39.0 & 8.72
 & 43.7 & 24.43
 & 3.0 & 5.73
 & 4.3 & 29.20 \\
\bottomrule
\end{tabular}%
}

\endgroup
\end{table}

\FloatBarrier

\subsection{Quality control and range expansion}
\label{sec:quality_control_results}

\paragraph{Quality control across iterations.}
We first test whether QCP can produce different levels of execution quality on request by evaluating it at each bin in $\{0,1,2,3\}$ at every iteration (Figure~\ref{fig:quality_control}). Even the initial policy trained only on demonstrations (iteration 0) follows the requested bins: on Square, for example, requesting the highest bin yields faster execution than the lowest, with mean steps to success of 193.7 at $q=3$ versus 308.1 at $q=0$, and the intermediate bins fall in between. We further find that this ordering holds for every attribute at every iteration, even though new data comes only from the highest bin. Iterative relabeling thus raises the quality of the highest bin over iterations while preserving the ability to select different execution behaviors from a single policy.

\begin{figure}[!ht]
    \centering
    \setlength{\parskip}{0pt}
    \setlength{\abovecaptionskip}{8pt}
    \setlength{\belowcaptionskip}{0pt}
    \includegraphics[width=0.90\linewidth]{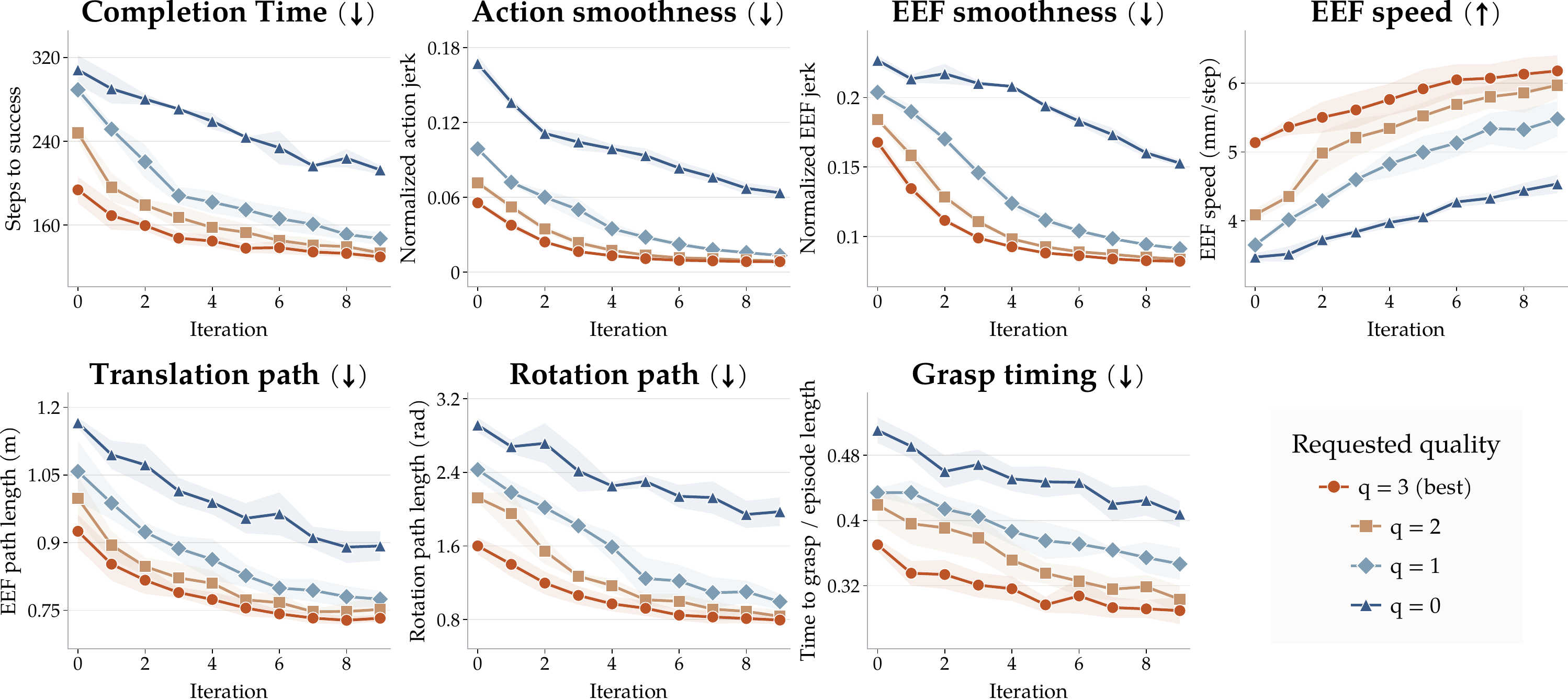}
    \par\nointerlineskip
    \caption{Quality achieved at each requested bin across iterations on Robomimic Square. The ordering of the bins is preserved while the whole range moves toward higher quality (arrows). Mean $\pm$ one standard deviation over three seeds.}
    \label{fig:quality_control}
\end{figure}

\paragraph{Quality beyond the initial demonstrations.}
Figure~\ref{fig:hqr_distribution_evolution} further shows that successful rollout distributions shift \textit{beyond the initial demonstration range} for Square smoothness and RoboEval joint path length. We quantify the distribution shift using the Wasserstein-1 distance between the initial demonstrations and final successful rollouts, normalized by the interquartile range (IQR) of the initial demonstration scores. This distance exceeds 1.5 for smoothness and joint path length, indicating large shifts from the initial demonstration distribution. For speed, the rollout distribution shifts toward shorter completion times, extending beyond the range observed in the initial demonstrations.\footnote{In our observations, the fastest rollouts appear to be close to the maximum physically achievable speed in this setup.}  These results show that online rollout collection and iterative learning enable QCP to expand its capable quality range beyond the initial demonstrations while successfully completing the task.

\begin{figure}[!htbp]
    \centering
    \setlength{\parskip}{0pt}
    \setlength{\abovecaptionskip}{7pt}
    \setlength{\belowcaptionskip}{0pt}
    \includegraphics[width=\linewidth]{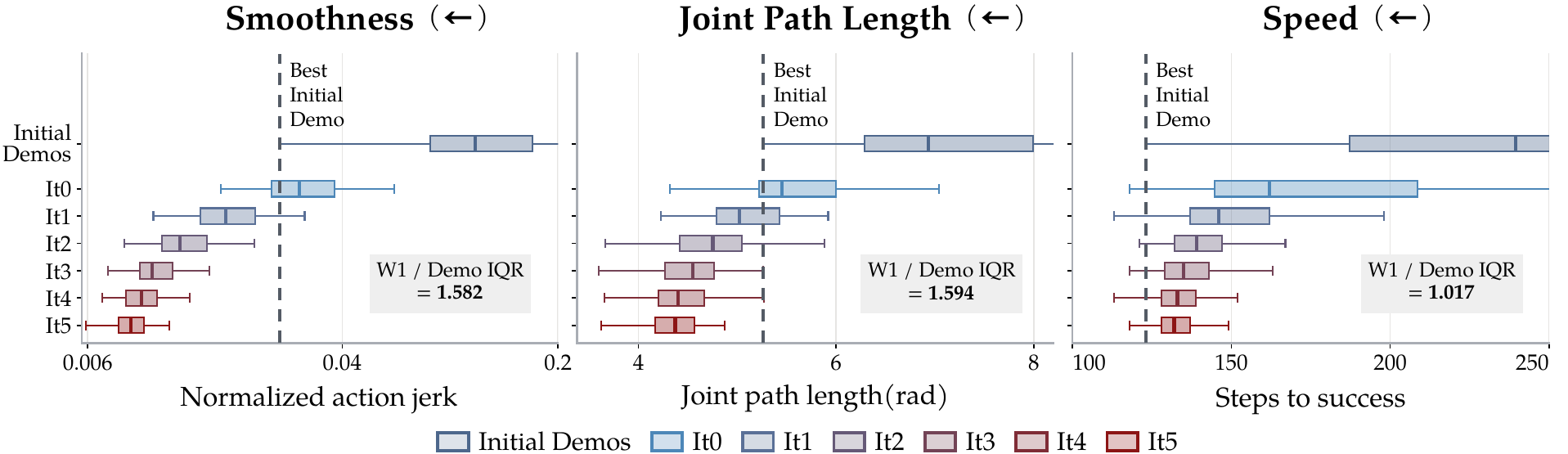}
    \par\nointerlineskip
    \caption{\textbf{Quality distributions across iterations.} Starting from the initial demonstration distribution, QCP shifts the quality distribution of successful rollouts toward higher quality, extending beyond the initial demonstration range. Arrows point toward higher quality.}
    \label{fig:hqr_distribution_evolution}
\end{figure}

\FloatBarrier

\subsection{Multi-attribute quality control and improvement}
\label{sec:multi_quality_results}

\begin{wrapfigure}{r}{0.40\textwidth}
    \centering
    \setlength{\parskip}{0pt}
    \setlength{\abovecaptionskip}{8pt}
    \setlength{\belowcaptionskip}{0pt}
    \includegraphics[width=\linewidth]{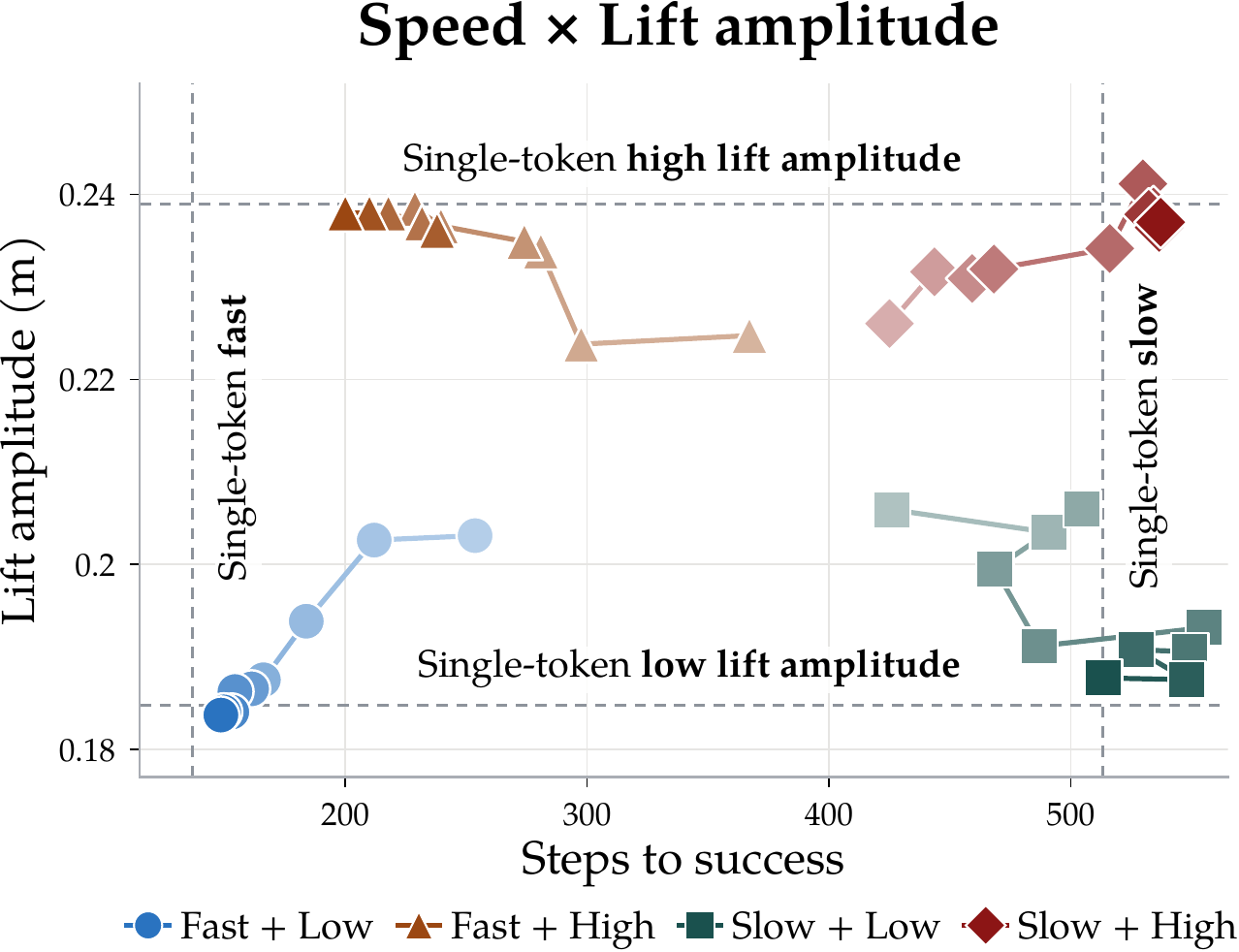}
    \par\nointerlineskip
    \caption{QCP jointly adjusts execution speed and lift amplitude in the requested directions over iterations.}
    \label{fig:joint_quality_control}
    \vspace{-8pt}
\end{wrapfigure}

We next evaluate whether QCP can control multiple quality attributes in the requested directions and improve them through iterative learning.

\paragraph{Bidirectional control of two quality attributes.}
On Robomimic Square, we train a separate QCP policy for each of the four combinations of faster or slower execution and smaller or larger lift amplitude.\footnote{Lift amplitude is the difference between the maximum and minimum end-effector height during execution.} Figure~\ref{fig:joint_quality_control} shows that both attributes change in the requested directions over iterations. Across all four combinations, the final policies achieve quality levels close to those of policies trained for each attribute alone, showing that QCP can pursue both objectives together with comparable quality.

\paragraph{Joint improvement across six attributes.}
We further evaluate whether QCP can jointly improve up to six quality attributes across four RoboEval tasks (Figure~\ref{fig:multi_quality_radar}). We condition each policy on six quality labels and request the best bin for every attribute during iterative learning (see Appendix~\ref{app:multi_quality_details} for experimental details). The final policies achieve better quality than initial BC and \textsc{BC}$_{\text{acc}}$ for most quality attributes across the evaluated tasks. These results suggest that QCP can jointly improve multiple aspects of execution even when a single policy is conditioned on up to six quality attributes.

\begin{figure}[!htbp]
    \centering
    \setlength{\parskip}{0pt}
    \setlength{\abovecaptionskip}{6pt}
    \includegraphics[width=\linewidth]{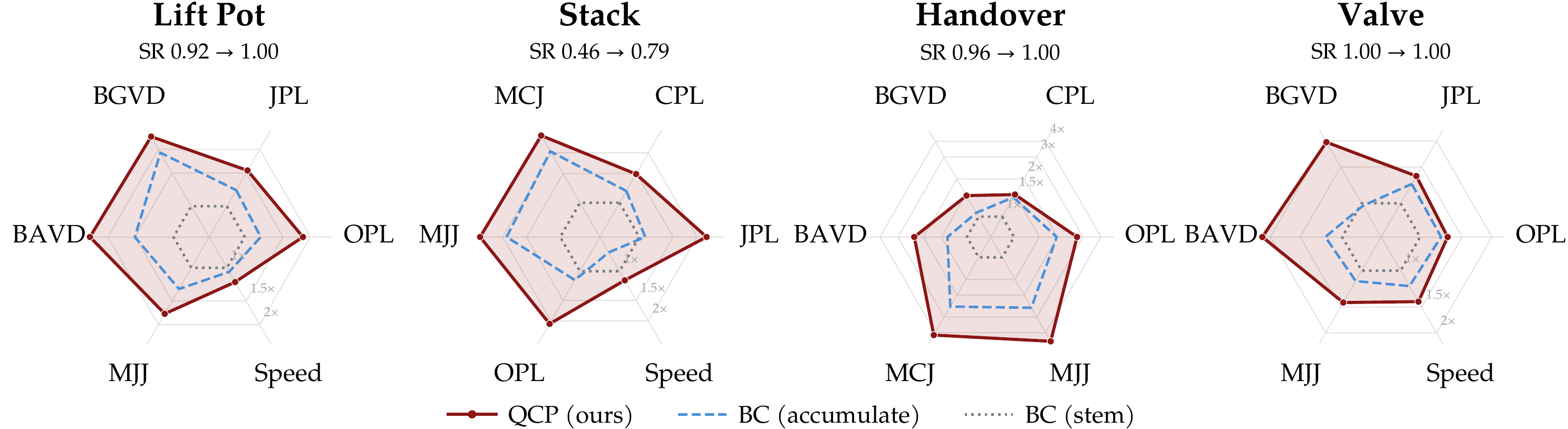}
    \par\nointerlineskip
    \caption{\textbf{Joint improvement of six quality attributes on RoboEval.} Final QCP and \textsc{BC}$_{\text{acc}}$ policies are compared with initial BC on four tasks. Each axis shows the initial BC score divided by the method's score, so the initial BC polygon lies on the $1\times$ ring and outward indicates better quality. Success rates of initial BC and final QCP are shown above each task.}
    \label{fig:multi_quality_radar}
\end{figure}

\FloatBarrier

\subsection{Quality control for real robot tasks}
\label{sec:generalization_results}

We evaluate whether controlling execution quality leads to practical gains in real-world robot tasks. We consider two tasks with distinct quality requirements: execution speed for the conveyor task, measured by end-effector speed, and transport stability for the wine-glass task, measured by the tray's tilt angle from horizontal (Appendix~\ref{app:quality_score_functions}).

\paragraph{Serving wine.}
The robot must carry a tray holding wine glasses from one table to another without dropping any glass. This task is challenging even for human operators: only 50 of 382 teleoperated attempts (13.1\%) succeeded during data collection, as keeping the tray level while moving proved difficult. Conditioning QCP on the highest stability bin reduces the mean tilt angle from 0.0660~rad for BC to 0.0482~rad, a 27.0\% reduction, while the success rate improves from 30.7\% to 42.0\% over iterations (Figure~\ref{fig:hqr_task_generalization}a). In contrast, \textsc{BC}$_{\text{filter}}$ improves tilt angle and success rate through iteration 1 but degrades at iteration 2, suggesting that retaining lower quality data provides broader coverage needed to sustain task success. Notably, even the initial QCP policy, trained only on demonstrations, already achieves a higher success rate than BC. We attribute this to the quality label, which separates stable from unstable executions in the demonstrations so that the policy does not average over them. The final policy succeeds far more often than teleoperation did during data collection.

\paragraph{Conveyor belt work.}
Twenty cans are placed on a conveyor belt moving at a fixed speed, one every three seconds, and throughput is defined as the number of cans successfully grasped per minute. Achieving high throughput under this fixed schedule requires both fast execution and a high success rate. Relative to BC, QCP increases mean end-effector speed from 188.6 to 257.6~mm/s and throughput from 8.2 to 13.0 cans per minute, improvements of 36.6\% and 59.3\% (Figure~\ref{fig:hqr_task_generalization}b). In contrast, \textsc{BC}$_{\text{filter}}$ falls behind QCP in end-effector speed after iteration 0 (217.7~mm/s at iteration 2), and its throughput drops from 8.3 to 6.3 cans per minute, below BC, consistent with the loss of coverage observed in wine serving. The initial QCP policy again already exceeds the throughput of BC, while the final QCP policy surpasses the 173.5~mm/s of human teleoperation, where high-speed control is challenging.

\begin{figure}[!htbp]
    \centering
    \includegraphics[width=\linewidth]{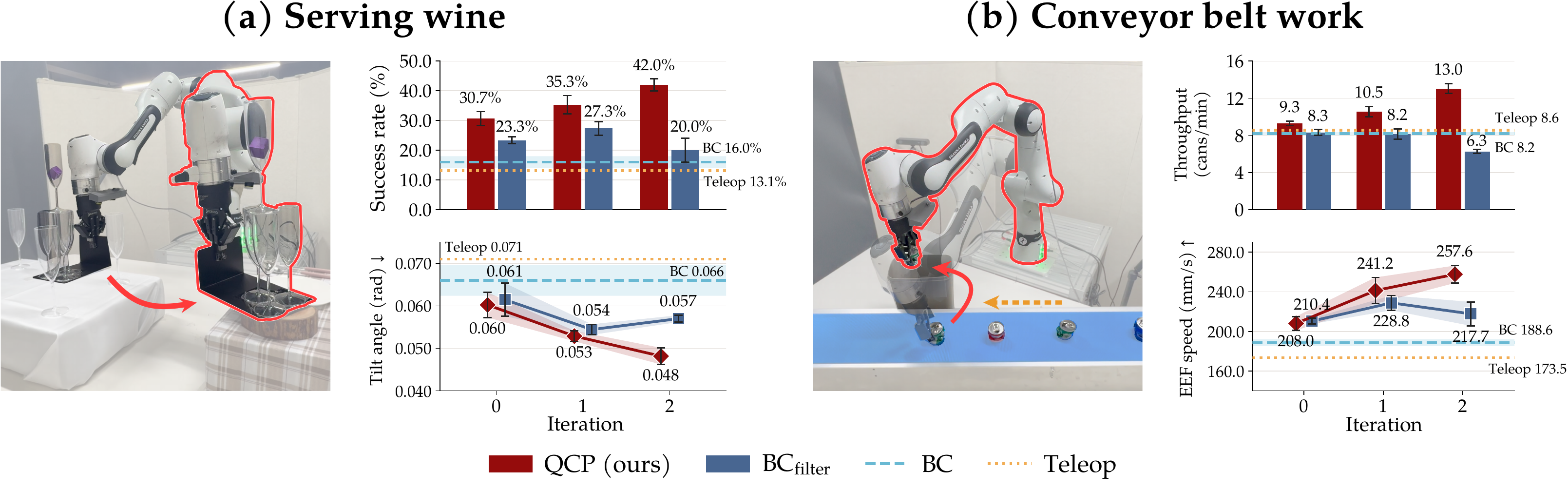}
    \vspace{-8pt}
    \caption{\textbf{Quality control on real robots.} QCP achieves higher task success than BC baselines while (a) reducing the tray tilt angle in wine serving and (b) increasing end-effector speed in conveyor work. Values are means over three seeds; error bars and the shaded band around BC denote $\pm$1 standard deviation.}
    \label{fig:hqr_task_generalization}
\end{figure}

\FloatBarrier
\section{Analysis}
\label{sec:analysis}

\begin{figure}[!t]
    \centering
    \begin{minipage}[b]{0.43\linewidth}
        \centering
        \includegraphics[width=\linewidth,height=4.2cm,keepaspectratio,trim={0 4.3 0 0},clip]{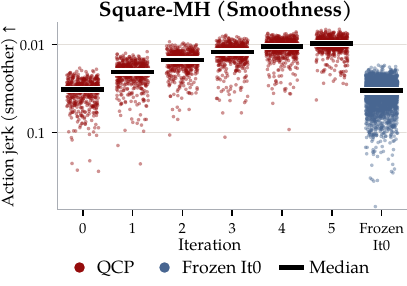}
    \end{minipage}\hfill
    \begin{minipage}[b]{0.53\linewidth}
        \centering
        \includegraphics[width=\linewidth,height=4.2cm,keepaspectratio]{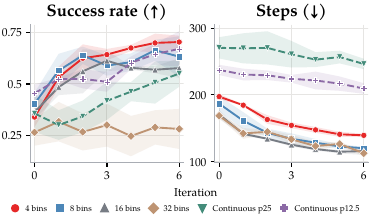}
    \end{minipage}
    \par
    \begin{minipage}[t]{0.43\linewidth}
        \caption{Iterative versus fixed policy rollouts at equal sample counts.}
        \label{fig:analysis_conditioning}
    \end{minipage}\hfill
    \begin{minipage}[t]{0.53\linewidth}
        \caption{Quantized labels with 4 to 32 bins versus continuous score conditioning on Square speed.}
        \label{fig:analysis_discretization}
    \end{minipage}
\end{figure}

\subsection{Effect of quality conditioning on rollout distributions}
\label{sec:analysis_conditioning}

We examine whether the distribution shift in Figure~\ref{fig:hqr_distribution_evolution} comes from the iterative hindsight relabeling loop or from quality conditioning alone, since a conditioned policy might reach rare behaviors simply by sampling more. To separate the two, we spend the same rollout budget in two ways under the best quality bin $q^\star=3$: 1,800 rollouts from the QCP policy frozen at iteration 0, or 300 rollouts at each of iterations 0 to 5 with retraining in between (Figure~\ref{fig:analysis_conditioning}). The fixed policy reproduces its iteration-0 distribution regardless of sample count, and none of its successful rollouts reaches the median smoothness of the final iteration. The improvement thus comes from policy updates on the relabeled data, which move the best bin to a quality level the previous policy could not produce.

\subsection{Continuous quality scores versus quantized quality labels}
\label{sec:analysis_discretization}

We examine why QCP uses quantized quality labels rather than continuous scores. With four quantile bins, roughly the best quarter of the training trajectories share the best label, giving the policy multiple examples of the requested behavior. Learning from multiple trajectories with the same label can provide a more stable training signal than conditioning on a score represented by only a few examples. The importance of training support is also evident in return conditioning, where Decision Transformer \citep{chen2021decisiontransformer} and RvS \citep{emmons2022rvs} can be unreliable for requests outside the training distribution \citep{brandfonbrener2022rcsl}. On Square, quantized quality labels execute faster than a fixed continuous target at every iteration, and four bins also reach higher task success from iteration 1 onward; increasing the number of bins beyond four buys a small gain in speed at a growing cost in task success (Figure~\ref{fig:analysis_discretization}). These results support simple quantization as an effective representation for quality control. Experimental details are provided in Appendix~\ref{app:analysis_details}.

\section{Conclusion}
\label{sec:conclusion}
We presented QCP, a robot policy that uses quantized quality labels to control individual execution attributes and their combinations at inference time. Iterative hindsight relabeling extends the controllable quality range beyond the initial demonstrations through online rollout collection and policy updates, without additional human annotation. Experiments on Robomimic and RoboEval show that QCP can improve execution quality and task success while retaining quality control across iterations. Real robot experiments further demonstrate that QCP can be directed to pursue the execution quality desired for a given task.

\bibliography{hqr}
\bibliographystyle{iclr2027_conference}

\clearpage
\FloatBarrier
\appendix
\section{Quality Score Functions}
\label{app:quality_score_functions}

Every quality attribute is measured on one complete trajectory, from the initial observation to the step at which the task is completed. Both simulators run at a control rate of 20\,Hz, so one step corresponds to 0.05\,s. Either direction of an attribute can be the preferred one depending on the task; for each attribute we state below the direction preferred in our experiments. The score $m(\tau)$ of Section~\ref{sec:quality_definition} is the measurement itself when a higher value is preferred and its reciprocal when a lower value is preferred, as in $m(\tau)=1/T$ for speed, so that the highest bin $q^\star=B-1$ always denotes the preferred end of the attribute; the figures and tables report the raw measurements.

\paragraph{Quantization.}
For every score, the bin edges are the $1/B,\ldots,(B-1)/B$ quantiles of the scores of the successful trajectories in the current dataset, recomputed at every iteration. A trajectory whose score coincides with an edge is assigned to the better bin. For the joint conditioning experiments, each attribute is quantized independently.

\paragraph{Robomimic.}
Let $T$ be the number of executed actions, $a_t\in\mathbb{R}^{6|\mathcal{A}|}$ the arm part of the action at step $t$ (the position and rotation deltas of every arm in $\mathcal{A}$, gripper dimensions excluded), $p_t^r\in\mathbb{R}^3$ the end-effector position of arm $r$, and $q_t^r$ its orientation quaternion. Let $s_t=\frac{1}{|\mathcal{A}|}\sum_{r}\lVert p_{t+1}^r-p_t^r\rVert_2$ denote the arm-averaged displacement per step. Square uses one arm and Transport two.
\begin{itemize}\setlength{\itemsep}{1pt}
\item \textbf{Speed (steps to success):} $T$, in steps. Lower is preferred in the main experiments; Section~\ref{sec:multi_quality_results} also requests slower execution.
\item \textbf{Action smoothness (normalized action jerk):} $\frac{1}{T-1}\sum_{t}\lVert a_{t+1}-a_t\rVert_2^2 \,/\, \frac{1}{T}\sum_t\lVert a_t\rVert_2^2$, dimensionless. Lower is better.
\item \textbf{End-effector smoothness (normalized EEF jerk):} $\frac{1}{|\mathcal{A}|(T-2)}\sum_{r,t}\lVert p_{t+2}^r-2p_{t+1}^r+p_t^r\rVert_2 \,/\, \frac{1}{T-1}\sum_t s_t$, dimensionless. Lower is better.
\item \textbf{End-effector speed:} $\frac{1}{T-1}\sum_t s_t$, in m per step. Higher is better.
\item \textbf{Translation path length:} $\frac{1}{|\mathcal{A}|}\sum_r\sum_t\lVert p_{t+1}^r-p_t^r\rVert_2$, in m. Lower is better.
\item \textbf{Rotation path length:} $\sum_t\frac{1}{|\mathcal{A}|}\sum_r 2\arccos\lvert\langle q_{t+1}^r,q_t^r\rangle\rvert$, in rad. Lower is better.
\item \textbf{Grasp timing:} the index of the first step with a closing gripper command divided by $T$, averaged over arms. Lower is better.
\item \textbf{Lift amplitude:} $\frac{1}{|\mathcal{A}|}\sum_r\bigl(\max_t z_t^r-\min_t z_t^r\bigr)$ with $z_t^r$ the height of the end-effector, in m. Section~\ref{sec:multi_quality_results} requests both a smaller and a larger lift amplitude.
\end{itemize}

\paragraph{RoboEval.}
The following scores use the metric implementation of RoboEval~\citep{wang2025roboeval}. Let $p_t^r$ be the wrist position of arm $r$, $q_t^r$ its orientation quaternion, and $\theta_t^r\in\mathbb{R}^7$ its joint configuration.
\begin{itemize}\setlength{\itemsep}{1pt}
\item \textbf{Completion time (CT, ``Speed''):} the number of control steps until success divided by 20\,Hz, in simulated seconds. Lower is better. RoboEval reports wall-clock time; we use simulated time so that the score does not depend on the machine.
\item \textbf{Joint path length (JPL):} $\sum_r\sum_t\lVert\theta_{t+1}^r-\theta_t^r\rVert_2$, in rad, summed over both arms. Lower is better.
\item \textbf{Cartesian path length (CPL):} $\sum_r\sum_t\lVert p_{t+1}^r-p_t^r\rVert_2$, in m, summed over both arms. Lower is better.
\item \textbf{Orientation path length (OPL):} $\sum_r\sum_t 2\arccos\lvert\langle q_{t+1}^r,q_t^r\rangle\rvert$, in rad, summed over both arms. Lower is better.
\item \textbf{Mean joint jerk (MJJ):} the mean over arms and steps of $\lVert\dddot{\theta}_t^r\rVert_2$, with velocity, acceleration, and jerk obtained by successive finite differences with $\Delta t$, in rad/s$^3$. Lower is better.
\item \textbf{Mean Cartesian jerk (MCJ):} the same construction on $p_t^r$, in m/s$^3$. Lower is better.
\item \textbf{Bimanual arm velocity difference (BAVD):} $\frac{1}{T}\sum_t\bigl\lvert\,\lVert\dot\theta_t^{L}\rVert_2-\lVert\dot\theta_t^{R}\rVert_2\bigr\rvert$ using the simulator joint velocities, in rad/s. Lower is better.
\item \textbf{Bimanual gripper vertical difference (BGVD):} $\frac{1}{T}\sum_t\lvert z_t^{L}-z_t^{R}\rvert$ with $z$ the wrist height, in m. Lower is better.
\end{itemize}

\paragraph{Real robot.}
Let $\mathbf{p}_t$ be the end-effector position at timestep $t$, $\mathbf{z}_t^{\text{eef}}$ the unit $z$-axis of the end-effector frame, $\mathbf{z}^{\text{world}}$ the unit $z$-axis of the world frame, and $T$ the episode length.
\begin{itemize}\setlength{\itemsep}{1pt}
\item \textbf{End-effector speed (conveyor work):}
\begin{equation}
    \bar{v} = \frac{1}{T} \sum_{t=1}^{T} \left\lVert \dot{\mathbf{p}}_t \right\rVert_2,
    \label{eq:real_speed}
\end{equation}
in mm/s, computed over successful grasp segments. Higher is better.
\item \textbf{Tilt angle (wine serving):}
\begin{equation}
    \bar{\theta} = \frac{1}{T} \sum_{t=1}^{T}
    \left|\arccos\!\left( \mathbf{z}_t^{\text{eef}} \cdot \mathbf{z}^{\text{world}} \right)\right|,
    \label{eq:real_tilt}
\end{equation}
in rad, computed over successful episodes. Since the tray is horizontal when it is rigidly held by an end-effector aligned with $\mathbf{z}^{\text{world}}$ (Figure~\ref{fig:hqr_task_generalization}a), $\bar{\theta}$ directly reflects the uprightness of the wine glasses. Lower is better.
\end{itemize}

\section{Training and Evaluation Details}
\label{app:training_details}

\paragraph{Policy and optimization.}
All simulation policies share one architecture and optimizer. The velocity network is an MLP with three layers of width 512 (expansion factor 1, no dropout) and a 128-dimensional flow-time embedding; a two-layer MLP encodes the observation history of $L=2$ steps. The quality label is mapped to a learned embedding and concatenated with the encoded observation before the velocity network. Policies are trained from scratch at every iteration with AdamW (learning rate $10^{-4}$, weight decay $10^{-5}$, gradient-norm clipping at 10), a cosine learning-rate schedule over the training steps, batch size 1024, and an exponential moving average of the weights with rate 0.995. Actions are sampled with nine Euler steps of the learned velocity field. Conditioned on an observation history of $L=2$ steps, the policy predicts a chunk of $H$ actions that starts at the first observation of the history; the $L-1$ actions aligned with past observations are discarded, the next $H-L$ actions are executed, and the last one is dropped before replanning. Both choices follow the original implementation of Diffusion Policy~\citep{chi2023diffusionpolicy}. The quality bins are recomputed as adaptive quartiles of the successful trajectories in the current dataset at every iteration, and rollouts are collected with the best bin requested. Table~\ref{tab:hparams} lists the settings that differ between experiments.

\begin{table}[!htbp]
\centering
\caption{\textbf{Per-experiment settings.} Steps are gradient steps per iteration. ``Number of iterations'' counts iterations including iteration 0, which is trained on the demonstrations alone. The action chunk is written as chunk length / executed steps.}
\label{tab:hparams}
\footnotesize
\setlength{\tabcolsep}{3pt}
\begin{tabular}{@{}lcccccccc@{}}
\toprule
 & \multicolumn{3}{c}{Main results (Figs.~\ref{fig:hqr_main_results}, \ref{fig:hqr_distribution_evolution}, Tab.~\ref{tab:rl_main})} & Fig.~\ref{fig:quality_control} & Fig.~\ref{fig:joint_quality_control} & Fig.~\ref{fig:multi_quality_radar} & Fig.~\ref{fig:analysis_conditioning} & Fig.~\ref{fig:analysis_discretization} \\
\cmidrule(lr){2-4}
 & Square & Transport & RoboEval & Square & Square & RoboEval & Square & Square \\
\midrule
Number of iterations & 6 & 6 & 6 & 10 & 10 & 7 & 6 & 7 \\
Steps per iteration & 100k & 100k & 30k & 30k & 30k & 30k & 30k & 30k \\
Episode limit (steps) & 400 & 700 & 400 & 400 & 800 & 400 & 800 & 400 \\
Action chunk & 6 / 4 & 10 / 8 & 10 / 8 & 10 / 8 & 10 / 8 & 10 / 8 & 10 / 8 & 10 / 8 \\
\bottomrule
\end{tabular}
\end{table}

\paragraph{Reported statistics.}
Success rates are computed over all rollouts of an evaluation batch. Quality is computed as the mean over the successful rollouts of the batch, first within each seed and then averaged across seeds with equal weight.

\paragraph{Real robot experiments.}
We use a Franka Research 3 arm with the DROID hardware and camera setup~\citep{khazatsky2024droid}. The policy is $\pi_{0.5}$~\citep{black2025pi05} fine-tuned with its original recipe for 10 epochs with batch size 32 at every iteration, starting from the pretrained checkpoint each time; the quality label is inserted into the language prompt as text. For wine serving, the initial dataset contains the 50 successful episodes among 382 teleoperated attempts; each iteration collects 50 rollouts, and success requires that the tray reaches the target table with no glass fallen. For conveyor work, the initial dataset contains 400 successful grasp segments from 936 teleoperated segments; each iteration collects 20 rollouts of 20 arriving cans, that is 400 segments, with cans arriving at one can every three seconds on a belt of fixed speed. Throughput is the success fraction multiplied by the arrival rate of 20 cans per minute. Speed is measured on successful segments and tilt on successful episodes. The quality label requested at collection is the best bin, $\textsc{BC}_{\text{filter}}$ retrains on the best 20\% of successful trajectories along the target axis at every iteration, and BC denotes $\pi_{0.5}$ fine-tuned on the initial dataset without a quality label.

\section{Baseline Details}
\label{app:baseline_details}

The BC baselines, SORL, and SCIQL use the network, optimizer, training steps, rollout budget, retraining from scratch, success filter, and seeds of QCP (Appendix~\ref{app:training_details}); they differ only in how the training set is formed and in the learning objective. DSRL and EXPO are described separately below.

\paragraph{$\textsc{BC}_{\text{acc}}$.}
$\textsc{BC}_{\text{acc}}$ runs the loop of Algorithm~\ref{alg:qcp} without a quality label: rollouts are collected by the unconditioned policy, successful ones are appended to the dataset, and a plain behavior-cloning policy is retrained on the accumulated set.

\paragraph{$\textsc{BC}_{\text{filter}}$.}
$\textsc{BC}_{\text{filter}}$ additionally ranks the successful trajectories of the accumulated set (demonstrations included) by the target score at every iteration and keeps the best 10\%. At every iteration the filter is applied again to the whole accumulated set, not to the previously filtered set, so the filtered training set grows with the accumulated data rather than shrinking over iterations; the setting does not force the dataset to become smaller. On Robomimic at least 20 trajectories are kept, and on RoboEval the filter is implemented as a sampling weight of 0 or 1 without a floor. In the real robot experiments the filter keeps the best 20\%, because the initial datasets are small (50 episodes for wine serving) and a 10\% filter would leave too few trajectories to train on.

\paragraph{SORL and SCIQL.}
Both methods proceed in two stages at every iteration. In the first stage, a value function is learned with implicit Q-learning~\citep{kostrikov2022iql} on a chunk-level MDP whose state is the two-step observation history, whose action is the executed action chunk, and whose discount is 0.99 per decision; the task reward is 1 on the chunk in which the task succeeds and 0 otherwise. The value and twin Q networks are two-layer MLPs of width 256 with layer normalization, trained for 60k steps with Adam (learning rate $3\times10^{-4}$, cosine schedule, batch 1024), expectile 0.7, and Polyak averaging with rate 0.005. Advantages are divided by the 99th percentile of their magnitude, and the resulting weights $\exp(\beta A)$ are clipped at 100.

In the second stage, the actor is trained. It is our quality-conditioned flow policy with the same network and the same number of gradient steps as QCP (100k on Robomimic, 30k on RoboEval), and the weights from the first stage are used as sampling probabilities over the training samples. The two methods differ in how the weights are formed. SORL~\citep{mao2024sorl} trains on the samples whose label equals the requested one, weighted by $\exp(\beta A^{r})$ with $A^{r}$ the task advantage. SCIQL~\citep{petitbois2026sciql} relabels every trajectory under each of the four labels, learns a style advantage $A^{\lambda}$ from the reward $\mathbf{1}\{z=q(\tau)\}$, and weights samples by $\exp(\beta\,\xi)$ with $\xi=A^{\lambda}+\sigma(A^{\lambda})A^{r}$, as in the original paper: the style advantage is the primary term, and the task advantage counts only for actions that already match the requested style. Both stages are retrained from scratch at every iteration on the accumulated success-filtered set, and rollouts are collected with the best label requested. The main results use $\beta=3$; Appendix~\ref{app:quality_reward_weight} reports the other values of $\beta$ and explains this choice.


\paragraph{DSRL and EXPO.}
Both methods start from a frozen flow matching BC policy trained on the demonstrations for 100k steps with the network of Appendix~\ref{app:training_details} and sampled with nine Euler steps. DSRL~\citep{wagenmaker2025dsrl} learns a soft actor-critic policy over the initial noise of that policy (update-to-data ratio 20, discount 0.999, batch 256, learning rate $3\times10^{-4}$, twin critics of three layers of width 2048 with layer normalization). EXPO~\citep{dong2025expo} learns an edit policy and a critic ensemble on top of the frozen policy (eight candidate actions per decision, edit scale 0.1, update-to-data ratio 5, discount 0.99, batch 256, an ensemble of ten critics with the minimum over two), with half of every batch drawn from the frozen policy's own 300 initial rollouts. Both act on chunks of 4 actions on Square and 8 elsewhere.

The reward has two parts, a task reward and a quality reward:
\begin{equation}
r_t=\mathbf{1}\{\text{task succeeds at step }t\}+\lambda\,r^{\text{q}}_t .
\end{equation}
The task reward is 1 at the step where the task succeeds, and the episode ends there. The quality reward $r^{\text{q}}_t$ is a small penalty given at every step, and $\lambda$ sets how much it counts relative to the task reward.

For speed, the penalty is the same at every step, $r^{\text{q}}_t=-1/L_{\text{ref}}$, where $L_{\text{ref}}$ is the average length of the demonstrations. Summed over an episode, the quality reward equals minus the episode length divided by $L_{\text{ref}}$, so a shorter episode earns a higher return and an episode as long as a typical demonstration loses about 1.

For smoothness and joint path length, the penalty at step $t$ depends on how much the motion at that step exceeds what is typical in the demonstrations:
\begin{equation}
r^{\text{q}}_t=-\frac{1}{L_{\text{ref}}}\Bigl(\mathrm{clip}\bigl(\tfrac{j_t-J_{\text{ref}}}{J_{\text{ref}}},\,-1,\,5\bigr)-\bar c\Bigr),
\end{equation}
where $j_t$ is the per-step motion measure ($\lVert a_t-a_{t-1}\rVert^2$ over the arm dimensions on Robomimic, the joint-space increment on RoboEval), $J_{\text{ref}}$ is its average over the demonstrations, the clipping keeps a single very jerky step from dominating the return, and $\bar c$ is the average of the clipped term over the demonstrations, so that a demonstration-like step receives zero penalty on average. Dividing by $L_{\text{ref}}$ puts the return on the same scale as the speed reward.

All constants are computed once from the successful demonstrations of each task and then kept fixed: $L_{\text{ref}}=240$, $J_{\text{ref}}=0.0354$, $\bar c=-0.253$ on Square; $L_{\text{ref}}=630$, $J_{\text{ref}}=0.0669$, $\bar c=-0.145$ on Transport; $L_{\text{ref}}=136$, $J_{\text{ref}}=0.141$, $\bar c=0$ on Pack Box; $L_{\text{ref}}=71$, $J_{\text{ref}}=0.137$, $\bar c=0$ on Pick Book. DSRL multiplies the whole reward by $L_{\text{ref}}/\max(1,\lambda)$ so that the size of its returns stays similar across values of $\lambda$. Table~\ref{tab:rl_main} uses $\lambda=4$; Appendix~\ref{app:quality_reward_weight} reports the other values.

The policy is evaluated on 100 episodes every 120k environment steps, which is the number of steps of 300 rollouts at the 400-step limit of Square and RoboEval, and the paper reports the sixth evaluation, that is 720k environment steps after the prefill. Transport uses the same interval.

\section{Effect of the Quality Coefficient}
\label{app:quality_reward_weight}

The RL baselines have one coefficient that sets how strongly the quality signal enters training: the advantage temperature $\beta$ of SORL and SCIQL and the quality reward weight $\lambda$ of DSRL and EXPO. To choose the values used in the main results, we swept these coefficients on Robomimic Square for both attributes with the settings of Appendix~\ref{app:baseline_details}, three seeds per setting, at the same endpoint as Table~\ref{tab:rl_main} (Table~\ref{tab:rl_sweep_robomimic}). For each method we selected the setting that gives a good balance of success rate and quality, $\beta=3$ and $\lambda=4$, and used it for the main results on all tasks.

\begin{table}[!htbp]
\centering
\caption{\textbf{Coefficient sweep on Robomimic Square.} Each cell is success rate (\%) / quality of successful episodes (steps to success for Speed, action jerk for Smoothness; lower is better), mean $\pm$ standard deviation over three seeds. Left: SORL and SCIQL over the advantage temperature $\beta$. Right: DSRL and EXPO over the quality reward weight $\lambda$. Bold rows are the settings used in the main results ($\beta=3$, $\lambda=4$) and reproduce Table~\ref{tab:rl_main}.}
\label{tab:rl_sweep_robomimic}
\vspace{12pt}
\scriptsize
\setlength{\tabcolsep}{2.5pt}
\renewcommand{\arraystretch}{1.1}
\begin{minipage}[t]{0.535\linewidth}
\centering
\begin{tabular}{@{}c cc cc@{}}
\toprule
 & \multicolumn{2}{c}{SORL} & \multicolumn{2}{c}{SCIQL} \\
\cmidrule(lr){2-3}\cmidrule(lr){4-5}
$\beta$ & Speed & Smoothness & Speed & Smoothness \\
\midrule
\shortstack{0\\~} & \shortstack{34.0$\pm$3.5 /\\213.8$\pm$17.0} & \shortstack{36.7$\pm$5.3 /\\0.0118$\pm$0.0010} & \shortstack{77.1$\pm$9.0 /\\251.2$\pm$25.6} & \shortstack{72.3$\pm$6.0 /\\0.0205$\pm$0.0017} \\
\addlinespace[2pt]
\shortstack{1\\~} & \shortstack{39.0$\pm$6.0 /\\206.5$\pm$18.8} & \shortstack{40.3$\pm$8.3 /\\0.0114$\pm$0.0012} & \shortstack{78.0$\pm$8.7 /\\231.0$\pm$26.7} & \shortstack{79.3$\pm$6.2 /\\0.0187$\pm$0.0018} \\
\addlinespace[2pt]
\shortstack{2\\~} & \shortstack{45.0$\pm$9.5 /\\189.4$\pm$9.8} & \shortstack{51.3$\pm$3.7 /\\0.0126$\pm$0.0007} & \shortstack{74.7$\pm$4.3 /\\218.9$\pm$18.6} & \shortstack{79.7$\pm$9.8 /\\0.0176$\pm$0.0018} \\
\addlinespace[2pt]
\shortstack{\textbf{3}\\~} & \shortstack{\textbf{53.2$\pm$7.8} /\\\textbf{156.0$\pm$18.1}} & \shortstack{\textbf{48.4$\pm$6.5} /\\\textbf{0.0115$\pm$0.0010}} & \shortstack{\textbf{68.7$\pm$5.2} /\\\textbf{206.5$\pm$23.2}} & \shortstack{\textbf{83.0$\pm$6.0} /\\\textbf{0.0128$\pm$0.0007}} \\
\addlinespace[2pt]
\shortstack{4\\~} & \shortstack{41.7$\pm$5.7 /\\188.1$\pm$18.0} & \shortstack{43.3$\pm$4.9 /\\0.0142$\pm$0.0008} & \shortstack{70.7$\pm$8.4 /\\212.5$\pm$14.6} & \shortstack{85.0$\pm$4.7 /\\0.0185$\pm$0.0022} \\
\addlinespace[2pt]
\shortstack{10\\~} & \shortstack{48.7$\pm$5.5 /\\194.2$\pm$23.3} & \shortstack{44.7$\pm$3.7 /\\0.0212$\pm$0.0017} & \shortstack{63.7$\pm$7.7 /\\208.2$\pm$19.0} & \shortstack{59.2$\pm$6.0 /\\0.0163$\pm$0.0012} \\
\addlinespace[2pt]
\shortstack{16\\~} & \shortstack{47.3$\pm$7.4 /\\159.5$\pm$18.0} & \shortstack{44.7$\pm$4.6 /\\0.0201$\pm$0.0015} & \shortstack{52.3$\pm$5.1 /\\195.3$\pm$13.8} & \shortstack{61.0$\pm$9.4 /\\0.0177$\pm$0.0014} \\
\addlinespace[2pt]
\bottomrule
\end{tabular}
\end{minipage}\hfill
\begin{minipage}[t]{0.455\linewidth}
\centering
\begin{tabular}{@{}c cc cc@{}}
\toprule
 & \multicolumn{2}{c}{DSRL} & \multicolumn{2}{c}{EXPO} \\
\cmidrule(lr){2-3}\cmidrule(lr){4-5}
$\lambda$ & Speed & Smoothness & Speed & Smoothness \\
\midrule
\shortstack{0\\~} & \shortstack{30.0$\pm$8.2 /\\196.2$\pm$15.6} & \shortstack{42.0$\pm$5.0 /\\0.0463$\pm$0.0055} & \shortstack{92.0$\pm$8.6 /\\161.8$\pm$16.4} & \shortstack{96.0$\pm$7.6 /\\0.1183$\pm$0.0116} \\
\addlinespace[2pt]
\shortstack{1\\~} & \shortstack{25.3$\pm$8.0 /\\192.2$\pm$17.4} & \shortstack{34.0$\pm$4.8 /\\0.0455$\pm$0.0039} & \shortstack{90.0$\pm$6.6 /\\164.2$\pm$13.8} & \shortstack{94.0$\pm$5.9 /\\0.1428$\pm$0.0159} \\
\addlinespace[2pt]
\shortstack{\textbf{4}\\~} & \shortstack{\textbf{37.0$\pm$4.2} /\\\textbf{208.8$\pm$22.4}} & \shortstack{\textbf{43.0$\pm$8.1} /\\\textbf{0.0469$\pm$0.0044}} & \shortstack{\textbf{96.0$\pm$4.1} /\\\textbf{162.5$\pm$10.7}} & \shortstack{\textbf{79.0$\pm$10.0} /\\\textbf{0.1574$\pm$0.0147}} \\
\addlinespace[2pt]
\shortstack{16\\~} & \shortstack{52.0$\pm$6.9 /\\224.4$\pm$20.1} & \shortstack{49.0$\pm$3.4 /\\0.0510$\pm$0.0034} & \shortstack{97.0$\pm$5.4 /\\163.8$\pm$14.6} & \shortstack{68.0$\pm$7.4 /\\0.1631$\pm$0.0176} \\
\addlinespace[2pt]
\bottomrule
\end{tabular}
\end{minipage}
\end{table}
\FloatBarrier

\section{Multi-Quality Conditioning Experiment Details}
\label{app:multi_quality_details}

\paragraph{Two-attribute control on Square.}
Figure~\ref{fig:joint_quality_control} uses Square with an 800-step episode limit so that slow executions can complete, ten policies, and three seeds. Each trajectory receives two labels, a speed quartile (steps to success) and a lift-amplitude quartile (Appendix~\ref{app:quality_score_functions}), both recomputed on the current dataset at every iteration. Each label is mapped through its own embedding and the two embeddings are concatenated with the encoded observation. A separate lineage is trained for each of the four corner commands, namely the lowest or highest speed quartile combined with the lowest or highest lift-amplitude quartile, and every lineage collects its rollouts under its own command. The dashed reference lines in Figure~\ref{fig:joint_quality_control} are the extreme iteration means of single-label lineages run with the identical protocol: fastest and slowest speed commands, and lowest and highest lift-amplitude commands. The joint commands reach the two single-label references on both coordinates within a few percent, at a lower success rate for the slow commands.

\paragraph{Six-attribute improvement on RoboEval.}
Figure~\ref{fig:multi_quality_radar} uses the state-based policy of Table~\ref{tab:hparams} on the Lift Pot, Stack Two Blocks, Handover, and Valve tasks of RoboEval with the demonstrations of the fixed-position variant, seven policies, 300 rollouts per iteration, and three seeds. Each trajectory receives six quartile labels, one per attribute, each mapped through its own embedding; all six labels are requested at their best bin during collection. The six attributes are chosen per task from the eight RoboEval metrics of Appendix~\ref{app:quality_score_functions} by dropping the two that a preliminary eight-attribute run showed to be unaffected by conditioning or to be in conflict with the task (Table~\ref{tab:six_axes}). The reference policies are the unconditioned BC policy trained on the demonstrations (initial BC) and the unconditioned loop that accumulates its successful rollouts for six iterations ($\textsc{BC}_{\text{acc}}$), both with the same trainer, demonstrations, and seeds. Per-seed values are means over successful rollouts of the evaluation batch and are averaged across seeds. Table~\ref{tab:six_axes_values} lists the values behind the radar plot.

\begin{table}[!htbp]
\centering
\caption{\textbf{Conditioned attributes per task.} The two dropped attributes and the reason.}
\label{tab:six_axes}
\vspace{12pt}
\small
\begin{tabular}{@{}llp{6.9cm}@{}}
\toprule
Task & Conditioned & Dropped (reason) \\
\midrule
Lift Pot & JPL, OPL, CT, MJJ, BAVD, BGVD & CPL (a fixed vertical lift leaves little Cartesian path to vary), MCJ (redundant with MJJ) \\
Stack & CPL, JPL, CT, OPL, MJJ, MCJ & BAVD (one-arm task, equal arm velocities are not a quality), BGVD (large by design of the task) \\
Handover & CPL, OPL, MJJ, MCJ, BAVD, BGVD & CT (in conflict with the smoothness attributes), JPL (redundant with CPL and OPL) \\
Valve & JPL, OPL, CT, MJJ, BAVD, BGVD & MCJ (nearly constant for a wrist rotation), CPL (path fixed by the valve geometry) \\
\bottomrule
\end{tabular}
\end{table}

\begin{table}[!htbp]
\centering
\caption{\textbf{Six-attribute quality on RoboEval.} Numerical values of Figure~\ref{fig:multi_quality_radar}: success rate and the six conditioned attributes at iteration 6 for QCP and $\textsc{BC}_{\text{acc}}$, and for the initial BC. Mean $\pm$ standard deviation over seeds; lower is better for every attribute; units as in Appendix~\ref{app:quality_score_functions}.}
\label{tab:six_axes_values}
\vspace{12pt}
\scriptsize
\setlength{\tabcolsep}{4pt}
\renewcommand{\arraystretch}{1.08}
\begin{tabular}{@{}lccccccc@{}}
\toprule
\textbf{Lift Pot} & SR & JPL & OPL & CT & MJJ & BAVD & BGVD \\
\midrule
Initial BC & 0.92$\pm$0.01 & 4.37$\pm$0.06 & 3.65$\pm$0.04 & 2.04$\pm$0.02 & 228.8$\pm$2.6 & 0.418$\pm$0.004 & 0.0052$\pm$0.0003 \\
$\textsc{BC}_{\text{acc}}$ & 0.99$\pm$0.00 & 3.59$\pm$0.07 & 3.08$\pm$0.06 & 1.95$\pm$0.03 & 176.6$\pm$2.6 & 0.278$\pm$0.015 & 0.0027$\pm$0.0003 \\
QCP & 1.00$\pm$0.00 & 2.82$\pm$0.09 & 1.98$\pm$0.09 & 1.72$\pm$0.00 & 130.5$\pm$6.4 & 0.173$\pm$0.019 & 0.0022$\pm$0.0002 \\
\midrule
\addlinespace[3pt]
\textbf{Stack} & SR & CPL & JPL & CT & OPL & MJJ & MCJ \\
\midrule
Initial BC & 0.46$\pm$0.02 & 1.98$\pm$0.03 & 8.16$\pm$0.24 & 6.64$\pm$0.05 & 5.72$\pm$0.11 & 153.2$\pm$3.0 & 39.2$\pm$0.6 \\
$\textsc{BC}_{\text{acc}}$ & 0.46$\pm$0.02 & 1.68$\pm$0.05 & 7.59$\pm$0.22 & 8.62$\pm$0.86 & 5.05$\pm$0.09 & 80.4$\pm$6.7 & 19.2$\pm$1.2 \\
QCP & 0.79$\pm$0.21 & 1.33$\pm$0.02 & 3.63$\pm$0.06 & 5.85$\pm$0.03 & 2.75$\pm$0.12 & 58.5$\pm$4.1 & 15.4$\pm$0.9 \\
\midrule
\addlinespace[3pt]
\textbf{Handover} & SR & CPL & OPL & MJJ & MCJ & BAVD & BGVD \\
\midrule
Initial BC & 0.96$\pm$0.00 & 1.10$\pm$0.09 & 2.24$\pm$0.29 & 169.6$\pm$10.6 & 48.1$\pm$3.2 & 0.293$\pm$0.004 & 0.0253$\pm$0.0008 \\
$\textsc{BC}_{\text{acc}}$ & 0.95$\pm$0.04 & 0.77$\pm$0.05 & 1.13$\pm$0.15 & 67.1$\pm$0.8 & 19.4$\pm$0.8 & 0.215$\pm$0.004 & 0.0236$\pm$0.0010 \\
QCP & 1.00$\pm$0.00 & 0.73$\pm$0.01 & 0.82$\pm$0.05 & 36.2$\pm$1.7 & 11.5$\pm$0.5 & 0.127$\pm$0.013 & 0.0172$\pm$0.0032 \\
\midrule
\addlinespace[3pt]
\textbf{Valve} & SR & JPL & OPL & CT & MJJ & BAVD & BGVD \\
\midrule
Initial BC & 1.00$\pm$0.00 & 1.98$\pm$0.11 & 1.68$\pm$0.08 & 2.03$\pm$0.08 & 62.8$\pm$0.4 & 0.486$\pm$0.023 & 0.0094$\pm$0.0004 \\
$\textsc{BC}_{\text{acc}}$ & 1.00$\pm$0.00 & 1.60$\pm$0.02 & 1.37$\pm$0.02 & 1.71$\pm$0.02 & 56.0$\pm$3.4 & 0.413$\pm$0.030 & 0.0096$\pm$0.0003 \\
QCP & 1.00$\pm$0.00 & 1.46$\pm$0.03 & 1.28$\pm$0.02 & 1.44$\pm$0.13 & 44.1$\pm$1.8 & 0.225$\pm$0.017 & 0.0048$\pm$0.0022 \\
\bottomrule
\end{tabular}
\end{table}

\begin{figure}[!htbp]
    \centering
    \includegraphics[width=0.7\linewidth]{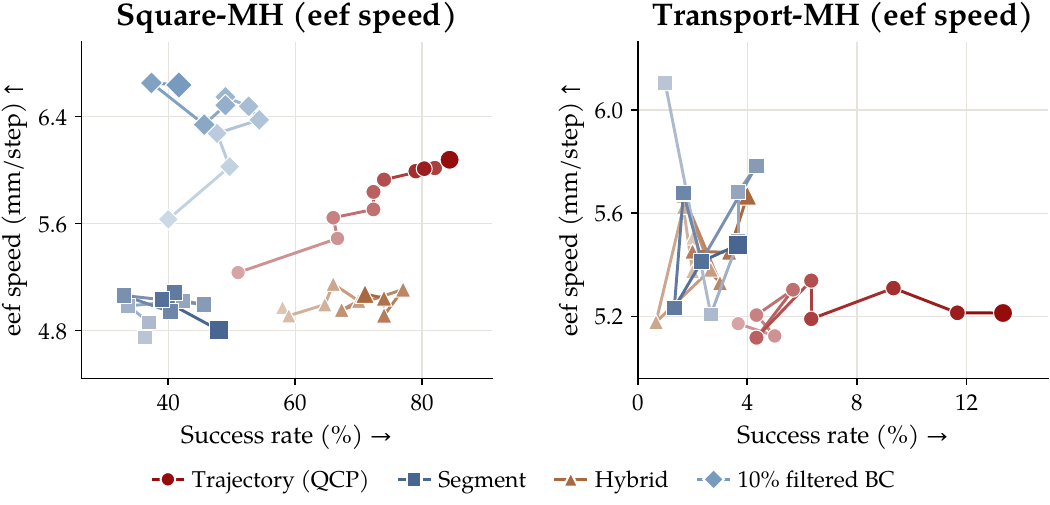}
    \caption{\textbf{Quality labels for shorter segments.} Success rate versus mean end-effector (EEF) displacement per step over successful rollouts on Square and Transport. We compare complete trajectory labels (QCP), segment labels, hybrid labels, and 10\% filtered BC over iterations 0--9; darker points indicate later iterations. Transport 10\% filtered BC has no successful rollouts, so its quality values are undefined and omitted.}
    \label{fig:analysis_subtrajectory}
\end{figure}

\section{Analysis Experiment Details}
\label{app:analysis_details}

\paragraph{Rollout collection comparison (Section~\ref{sec:analysis_conditioning}).}
The comparison uses the Square smoothness attribute with 30k training steps per iteration, an 800-step episode limit, and three seeds. The frozen variant draws 1,800 rollouts per seed from the checkpoint trained at iteration 0, conditioned on the best bin $q^\star=3$, with the dataset normalization that checkpoint was trained under, and without any further training. The iterative variant is the loop itself, which draws 300 rollouts per seed at each of iterations 0 to 5 under the same best-bin request, so both variants spend 1,800 rollouts per seed. The purpose of this comparison is to separate the effect of the quality-conditioned architecture from the effect of the iterative hindsight relabeling loop. The frozen variant has the architecture but no further loop iterations, so any additional shift of the iterative variant comes from the loop rather than from conditioning or sampling alone. Statistics are computed over successful rollouts pooled across seeds; per-seed medians agree with the pooled values to within 0.002. The frozen policy has a success rate of 54.5\% and a median action jerk of 0.0335 (interquartile range 0.028 to 0.043) over 2,943 successful rollouts, against 57.0\% and 0.0324 at iteration 0 of the loop. By iteration 5 the loop reaches 73.9\% and a median of 0.0098 (interquartile range 0.0088 to 0.0109). The smoothest frozen rollout has a jerk of 0.0171, above the 75th percentile of the iteration-5 distribution. The later iterations therefore produce behavior that the iteration-0 policy does not generate at all, not merely behavior that it generates rarely: the quality-conditioned architecture alone does not reach this range, and the additional shift comes from the iterative hindsight relabeling loop.

\paragraph{Quality representation (Section~\ref{sec:analysis_discretization}).}
This experiment examines how the quality condition should be represented: as a quantized label, and if so with how many bins, or as a continuous score. Figure~\ref{fig:analysis_discretization} compares six variants on the Square speed attribute, all run with the same loop (30k training steps and 300 rollouts per iteration, seven policies, three seeds). Four variants use quantized labels with $B=4$, 8, 16, or 32 equal-frequency bins, recomputed at every iteration, and always request the best bin; with more bins the requested bin covers a smaller share ($1/B$) of the training data. Two variants use a continuous score instead of a label: the score is normalized so that the best and worst initial demonstrations map to 0 and 1, this normalization is fixed for the whole loop, and the requested value is set once to the 25th or the 12.5th percentile of the initial demonstrations. Curves are means over seeds and bands are one standard deviation. At iteration 6, the four-, eight-, sixteen-, and thirty-two-bin variants reach 70.3\%, 63.2\%, 57.7\%, and 28.1\% success with 139, 120, 116, and 112 steps; the two continuous variants reach 55.0\% and 67.0\% success with 246 and 209 steps.

More bins make executions slightly faster but cost success, sharply so beyond sixteen bins, and neither continuous target matches the bins in speed at any iteration; in success, four bins exceed both continuous targets from iteration 1 onward. Because the continuous variants also use a different way of selecting the requested value, this experiment supports quartile labels with per-iteration recomputation as a reliable default rather than ruling out continuous conditioning in general.

\FloatBarrier
\section{Quality Labels for Shorter Segments}
\label{sec:analysis_subtrajectory}

This experiment tests whether a quality label assigned to shorter segments of a trajectory is a better training signal than one label for the complete trajectory. We compare three labeling options. \emph{Trajectory} is QCP as used in the main experiments: one label per trajectory. \emph{Segment} assigns a label to every decision step from the score measured in a window of $\pm 8$ steps around it. \emph{Hybrid} conditions on both the trajectory label and the segment label through two embeddings. A segment score depends strongly on the stage of the task (for example, approaching versus inserting), so for the segment labels the quartile boundaries are computed separately within each of ten intervals of normalized trajectory time, and the best quartile is requested at every step. All options run the same loop otherwise.

Figure~\ref{fig:analysis_subtrajectory} shows the comparison on the end-effector speed attribute for Square and Transport, with 10\% filtered BC as a reference, over ten iterations with 100k training steps and 300 rollouts per iteration and one seed. At iteration 9 on Square, the trajectory label reaches 84.3\% success at 6.08\,mm per step, the segment label 48.0\% at 4.80, and the hybrid 71.0\% at 5.07; on Transport, the trajectory label reaches 13.3\% success against 3.7\% and 4.0\% for the segment and hybrid labels, at similar speeds (5.2 to 5.7\,mm per step). The trajectory label therefore gives the largest gains in both success and speed on Square and the highest success on Transport. A segment label provides a more local signal, but the signal reflects the stage of the task as much as the quality of execution even after the stage-wise normalization, and the policy can improve a local score without making progress toward completion. These results support labeling complete trajectories; using finer quality signals may require methods that separate quality from task progress, which we leave to future work.

\end{document}